\documentclass[11pt]{article}
\usepackage{acl2015}
\usepackage{times}
\usepackage{url}
\usepackage{latexsym}
\usepackage{booktabs}
\usepackage{graphicx}
\usepackage{hyperref}

\title{The Influence of Data Pre-processing and Post-processing on Long Document Summarization}

\author{
  Kailun Dong,
  Xinwei Du,
  Yuchen Zhang,
  Yongsheng Li,
  Ruei-Yu Tsay
  }

\date{\today}

\begin{document}
\maketitle
\begin{abstract}
Long document summarization is an important and hard task in the field of natural language processing.
A good performance of the long document summarization reveals the model has a decent understanding of the human language.
Currently, most researches focus on how to modify the attention mechanism of the transformer to achieve a higher ROUGE score.
The study of data pre-processing and post-processing are relatively few.
In this paper, we use two pre-processing methods and a post-processing method and analyze the effect of these methods on various long document summarization models.
\footnote{You can find our implementation on Github: \url{https://github.com/Anthonyive/csci-544-project}. Our video demo:\url{https://www.youtube.com/watch?v=oVIVtOPeWEs}}

\end{abstract}

\section{Introduction}

Long document summarization is a hard and important task which requires the model to identify and extract the important information in the document and generate a fluent summary. 
A good performance in the long document summarization usually shows the model's decent understanding of natural language.

Long document summarization is a sequence to sequence task, and a general way to handle this task is by using an LSTM encoder and decoder with the attention mechanism.
Discourse-Aware \cite{cohan2018discourse} is a pioneering paper in the long document summarization, 
they provided two standard datasets (arXiv and pubMed) and proposed a hierarchical encoder and a discourse-aware decoder to generate the summary.
They also adopted copying mechanism to address the problem of unknown tokens and used a decoder coverage vector to avoid repeated phrases in the summary.
TLM model \cite{subramanian2019extractive} combined both extractive and abstractive methods and used a transformer model \cite{vaswani2017attention} to generate the summary.
T5 \cite{raffel2019exploring} is a language model which explored the transfer learning techniques and introduced a unified text-to-text framework.
BigBird \cite{zaheer2021big} proposed a sparse attention mechanism and reduced the attention complexity from quadratic to linear.
Longformer and its variant LED \cite{beltagy2020longformer} further modified the attention mechanism and combined local windowed attention with global attention.
HEPOS \cite{huang2021efficient} proposed a novel efficient encoder-decoder attention and achieved the state-of-the-art results on PubMed dataset.

Nowadays, more and more long document summarization researches focus on how to improve the attention mechanism of the transformer.
However, other research areas, like image retrieval, for example, have post-processing methods like PQ \cite{jegou2011aggregating}, DBA \cite{philbin2007object}, and QE \cite{5995601}.
These post-processing methods can further improve the performance of the model and the implementation of the above-mentioned post-processing methods are independent of the core image retrieval algorithms.
Whereas, decent pre-processing and post-processing methods are lacking in the area of long document summarization.

In this paper, we use two pre-processing methods and a post-processing method and analyze the effect of these methods.

\begin{figure*}[htbp]
\centerline{\includegraphics[width=0.9\textwidth]{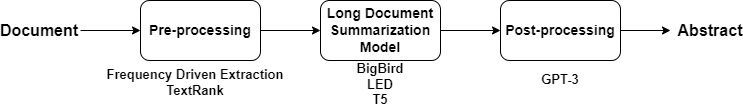}}
\caption{Long Document Summarization Pipeline.}
\label{fig1}
\end{figure*}

\section{Methods}
Figure \ref{fig1} shows our long document summarization pipeline.
For a given document, we first use a pre-processing method to extract important information from the original document.
Later, we feed the extracted text to a long document summarization model and generate the summary.
Then, we use a post-processing method to refine the summary generated by the model and form our final abstract.

\subsection{Pre-processing}
We use extraction-based methods to fulfill data pre-processing.
The extraction-based method is using some ranking algorithms to extract important sentences from original documents. 
Here in this paper, we used two different approaches of extractive methods to tackle the summarization problem. 

\subsubsection{Frequency Driven Extraction}\label{section_fde}

We first tried a Frequency Driven Approach. This method is useful for several reasons: 
\textbf{(1)} It gives us a baseline of what the target rouge-1, rouge-2, rouge-L scores are going to be; 
\textbf{(2)} It could become the first processing step for later steps; 
\textbf{(3)} Its fast, reliable, and light processing step makes it easy to pipe to other methods. 

Here we used a very simple scoring algorithm: 
Given a document $D$ with $N$ sentences, $D = \{S_1, S_2, ..., S_N\}$, we first remove stopwords and assign each remained word $w_i$ with equal weight 1
(we can also give values other than 1 based on prior knowledge, we set all weight to 1 for simplicity). 
Then, we calculate the value of each word $v_{w_i}$ by calculating the total weight of word $w_i$ in this document 
(in this case, we try to calculate the frequency of each word in this document since we set all weight to 1). 
\begin{equation}
v_{w_i} = count(D, w_i)
\label{eq1}
\end{equation}
For each sentence $S_i$, the value of each sentence $V_{S_i}$ euqals to the sum of the value of the words in the sentence.
\begin{equation}
% \sum_{w_i \in S_i, w_i \notin stopwords}^{} V_{w_i}
V_{S_i} = \sum_{w_i \in S_i}^{} V_{w_i}
\label{eq2}
\end{equation}
Finally, we choose the top $M$ sentences to create the summary. $M$ is a hyperparameter.
And we set $M$ as the nnumber of sentences of the ground true summary.

\begin{equation}
Result = top\_M(D)
\label{eq3}
\end{equation}

We acknowledge that the $top\_M$ approach might be limited as $M$ is dependent on the reference summary data. 
However, we are interested in finding the differences between whether using this frequency driven approach will increase the train and dev data accuracies or not. 

\begin{figure*}[htbp]
\centerline{\includegraphics[width=0.9\textwidth]{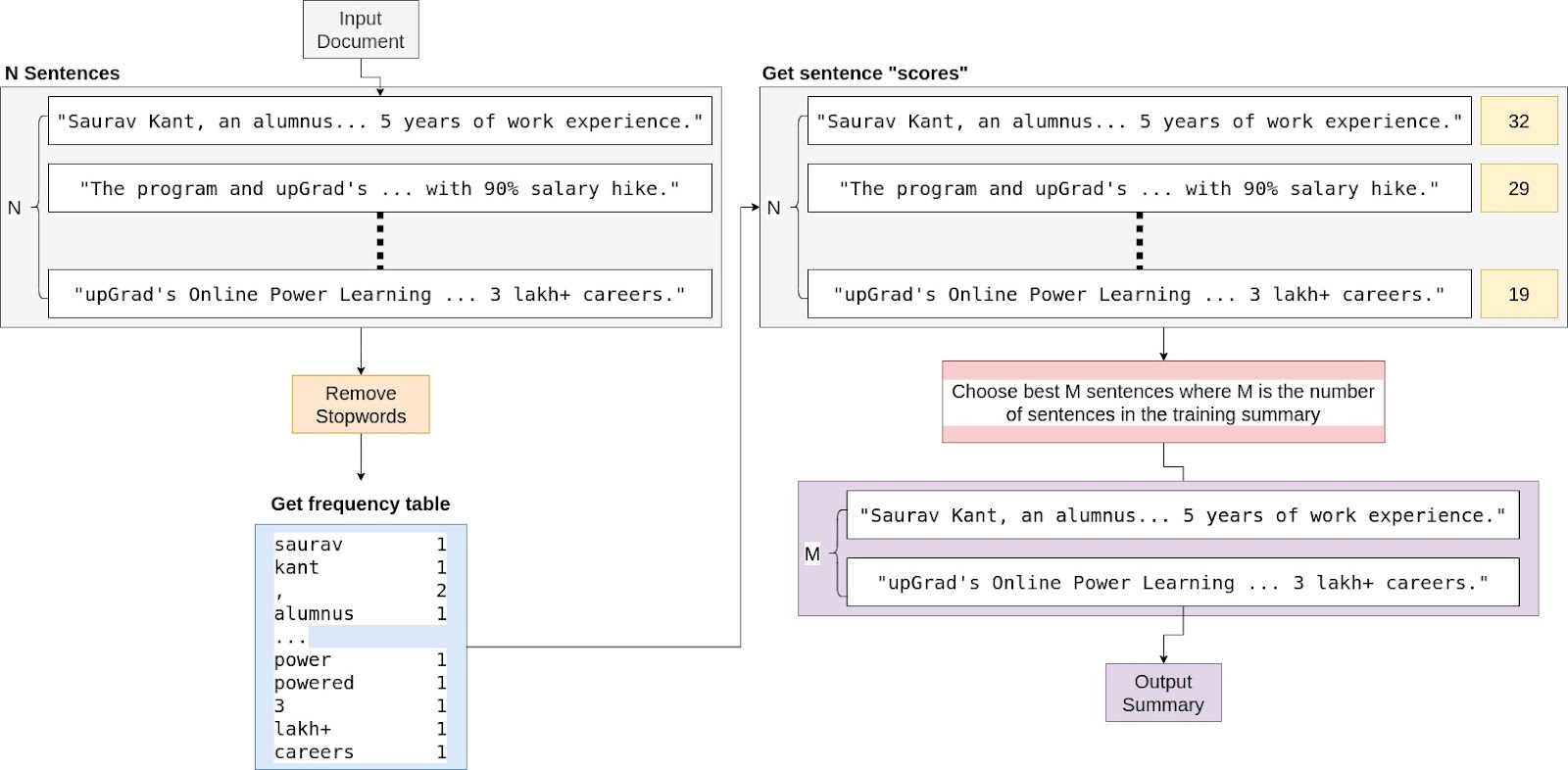}}
\caption{The Flowchart of Frequency Driven Algorithm.}
\label{fig2}
\end{figure*}

Figure \ref{fig2} show the flowchart of frequency driven algorithm.

\subsubsection{TextRank Algorithm}

The choosing of $M$ could be a problem for the algorithm in section \ref{section_fde} when we try to summary new data. 
To resolve finding $M$, we choose two approaches: 
One is we treat this $M$ as a fixed hyperparameter, 
i.e. a fixed number of sentences to find in the predicted summary. 
Another one is to use another approach without using M.

We use a graph-based method, TextRank \cite{mihalcea2004textrank}, which was influenced by Google’s Page Rank Algorithm \cite{page1999pagerank}, 
to do extractive document summarization.

\subsection{Long Document Summarization Models}
We adopt various long document summarization model to perform the summarization.
We use finetuned BigBird and LED models, and trained a T5 model from scratch.

A benefit of our summarization pipeline is that the implementation of long document summarization won't affect the data pre-processing step and post-processing step.

\subsection{Post-processing}

We use GPT-3 model to handle the post-processing step.
GPT-3 model is a transformer-based language model and is trained on large datasets which contain a huge amount of information about the reading text for many kinds of areas. 
Lots of experiments show that GPT-3 has a strong language generating ability.
Therefore, in our experiment, we use the GPT-3 model to do the fine post-processing step. 

Given the summary proposed by a long document summarization model, we use prompt as \textit{"Original [summary], Polished Sentence:"}. 
GPT-3 model will complete this prompt and generate the polished sentences.
We adopt the curie engine of GPT-3 which has a decent computing speed and performance.

\section{Experiments and Results}

\subsection{Datasets}
We used two datasets, namely arXiv and PubMed, to train our models and evaluate our models’ performances. 

The arXiv dataset and the PubMed dataset are both free and open-access resources for research use. 
Because the original full-size datasets are too large, the arXiv and PubMed datasets we used are subsets of their original data that take storage space over 1T and 90G. 
The arXiv dataset and PubMed dataset consist of 215913 and 133215 published scholarly articles respectively,
each having been separated into a training set, a validation set, and a testing set with similar distributions on articles’ sentence counts and token counts. 
The datasets also contain section information and citation information of the articles, 
while our team mainly focuses on the article texts to extract the articles’ summarizations.

Table \ref{tab1} shows the distribution characteristics of the two datasets.

\begin{table*}[htbp]
\caption{Dataset Distribution Characteristics.}
\begin{center}
\begin{tabular}{@{}lllll@{}}
\toprule
Dataset   & Num of doc & Avg num of sen & Avg num of tokens (doc) & Avg num of tokens (sen)\\ \midrule
arXiv  & 215k & 206 & 6029 & 29\\
% - training set  & 203k & 206 & 5905 & 29\\
% - validation set  & 64k & 204 & 5894 & 29\\
% - testing set  & 64k & 205 & 5905 & 29\\ 
\midrule
PubMed & 133k & 86 & 3048 & 33\\
% - training set  & 120k & 86 & 3043 & 33\\
% - validation set  & 66k & 87 & 3111 & 34\\
% - testing set  & 67k & 87 & 3092 & 34\\
\bottomrule
\end{tabular}
\end{center}   
\label{tab1}
\end{table*}

\begin{table*}[htbp]
  \caption{Summarization ROUGE score for long documents.}
  \begin{center}
  \begin{tabular}{@{}lllllll@{}}
  \toprule
                                                & \multicolumn{3}{c}{arXiv}    & \multicolumn{3}{c}{PubMed}         \\ \cmidrule(l){2-4} \cmidrule(l){5-7} 
  Model                                         & \multicolumn{1}{c}{R-1}  & \multicolumn{1}{c}{R-2}    & \multicolumn{1}{c}{R-L}  & \multicolumn{1}{c}{R-1}  & \multicolumn{1}{c}{R-2}   & \multicolumn{1}{c}{R-L}   \\ \midrule
  % BigBird                                       &  41.83  &  17.60  & 33.80  &  44.08  &  19.60   &  36.10  \\ %100
  BigBird                                       &  \textbf{41.94}  &  \textbf{21.59}  & \textbf{36.56}  &  \textbf{42.45}  &  18.18   &  \textbf{37.74}  \\
  + Extraction                                 & 27.54 & 7.63 & 17.06 & 34.21 &10.89 & 20.65 \\
  + TextRank                                  & 35.58 & 10.36 & 19.59 &  42.15  &  \textbf{27.80}  &  32.95 \\
  + GPT-3                                       &  42.16 & 21.86  & 36.75 & 34.95  &  12.73 &  31.07  \\\midrule
  % LED                                           &  44.56  &  19.90   &  39.57  & 46.33    & 20.22      & 40.51  \\ %100
  LED                                           &  \textbf{45.96}  &  \textbf{26.97}   &  \textbf{42.48}  & 43.10    & 17.11      & \textbf{32.76}  \\
  + Extraction                                 & 37.13  & 12.46  & 28.01  & 41.56 &18.95 & 24.02 \\
  + TextRank                                  & 40.43  &  11.23 & 25.00 & \textbf{49.53}  &  \textbf{21.45} &  29.48 \\
  + GPT-3                                       & 38.22  & 21.46  & 35.67 & 38.39  & 14.40  & 27.68 \\\midrule
  T5 (Fail)                                           &  12.65  &  5.45  & 10.82  &  14.42  &  5.58   &  12.99  \\
  % GPT-3 (Only)                                  &  46.45  &  37.35  & 40.80  &  39.02  &  27.30   &  31.94  \\
  Extraction (Only)                                   &  27.05  &  7.94  & 13.87  &  34.65  &  12.40   &  19.66  \\
  TextRank (Only)                                     & 29.49 & 9.58 & 14.86 &  31.34  &  12.25  & 17.35    \\\bottomrule
  \end{tabular}
  \end{center}   
  \label{tab2}
  \end{table*}
\subsection{Model Setting}
In order to better test the generalization ability of our proposed method, we use a variety of long document models.
The following describes how we use these models.

~\\
\noindent\textbf{BigBird:} 
We used two BigBird \cite{zaheer2021big} models fintuned on arXiv and PubMed datasets respectly to obtain the Rouge performance of BigBird model on these two datasets.
The maximum number of input tokens for BigBird is 4096. 
To generate the summary, we set the length penalty as 0.8, number of beam search as 5 and the maximum number of output tokens as 256.

~\\
\noindent\textbf{LED:} 
We used Longformer-Encoder-Decoder (LED), a variant of Longformer model \cite{beltagy2020longformer}, designed for long document summarization.
We used two fintuned LED models pretrained on arXiv and PubMed datasets respectly. 
The maximum number of input tokens of LED is 16384, but because of the limitations of GPU memory, 
we set the maximum number of input tokens as 8192. 
Longformer proposed a global attention mechanism, which can help the model to understand the documents better.
We put gloabl attention on the start token of each sentence.

~\\
\noindent\textbf{T5:} 
T5 \cite{raffel2019exploring} is a unified Text-to-Text Transfer Transformer pre-trained on a large text corpus and has been demostrated
to achieve the state-of-the-art performance on many NLP tasks. 
We fine-tuned long document summarization on the T5 pre-trained model using PubMed and Arxiv dataset, 
the maximum input length is set to 4096 and the maximum generation length is set to 256 due to GPU memory limits.

\subsection{Results and Discussion}
Table \ref{tab2} shows the Rouge score of the BigBird model and LED model, 
and the score when we apply pre-processing and post-processing methods. 
(Due to the limitation of computing resources and time, 
We only tested the first 100 test data in each dataset. 
Therefore, instead of focusing on the specific Rouge value, 
please focus on the difference in Rouge between different methods.)
We also listed the performance of our T5 model, which is trained from scratch, 
and the performance when we only use the pre-processing methods. 

According to Table \ref{tab2}, pre-processing method TextRank may increase the performance sometimes and decreases the performance at other time.
Pre-processing method finds the important sentence. It will help long document summarization models discard unimportant content and reduce distractions.
However, the pre-processing methods are not useful all the time, when it extracts unimportant content and discard important sentence by mistake, this method will decrease the summary performance.

As for post-processing, when we try to apply GPT-3 as a post-processing method, the Rouge score of the model decreases.
It makes sense since we only give GPT-3 model the summary generated by the long document summarization model. 
We try to input the original whole paper but it is too long to be handled by GPT-3.

Moreover, T5 does not perform well on long document summarizations according to the Rouge score, 
though it claimed to get good performance in summarization in the original T5 paper on short documents of CNN news. 
We think this may be because that the average length of CNN news is only 656 will the average length of arXiv and Pubmed datasets are more than 3000.
In addition, the abstractive summary generated by the T5 model is very short, 
the original summary of PubMed has a median length of 226 and only has a median length of 10 for generated summary.

\section{Conclusions and Future Work}
To explore how to further improve the performance of the long document summarization, 
we used frequency driven extracted algorithm and TextRank algorithm for data pre-processing,
and use GPT-3 for post-processing. The experiment results show that our proposed method can not improve the performance of the model.

In future work, we need to why study will T5 model generate such short summaries and if there are ways to modify it to adapt to long document summarization. 
We also want to try larger models such as T5-base and T5-large to see if we can improve the performance.

Moreover, we believe a decent post-processing method is needed not only for long document summarization, 
but also for other language generation tasks such as machine translation, dialogue generation.
A model dedicated to polishing articles may be useful for NLP tasks.

\bibliographystyle{acl}
\bibliography{ref}

\end{document}